\documentclass[review]{elsarticle}
\graphicspath{ {./images/} }
\usepackage{graphicx}
\usepackage{float}
\usepackage{resizegather}
\usepackage{hyperref}
\usepackage{float}
\usepackage{verbatim} 
\usepackage{apalike}
\restylefloat{figure}
\usepackage{subcaption}
\floatstyle{plaintop} 
\restylefloat{table}
\usepackage{natbib}
\usepackage{amsmath,amssymb,amsfonts}
\usepackage{algorithm}
\usepackage{algpseudocode}
\usepackage{xcolor}
\usepackage{array}
\usepackage{multirow}
\usepackage{multicol}
\usepackage{tabularray}
\usepackage{longtable}
\usepackage{xcolor}

\makeatletter
\def\nonumnote{\gdef\@thefnmark{}\@footnotetext}
\makeatother

\journal{Computer Vision and Image Understanding}

\begin{document}
	\begin{frontmatter}
		
		
		\begin{titlepage}
			\begin{center}
				\vspace*{1cm}
				
				\textbf{Are Candidate Models Really Needed for Active Learning?}
				
				\vspace{1.5cm}
		\nonumnote{Accepted for publication in Computer Vision and Image Understanding (CVIU).}		
				Harshini Mridula Mohan$^{a}$ (harshinimm.btech22@rvu.edu.in), Maanya Manjunath$^a$ (maanyam.btech22@rvu.edu.in), Vipul Arya$^a$ (krishnamva.btech22@rvu.edu.in), S.H. Shabbeer Basha$^b$ (shabbeer.basha@vidyashilp.edu.in), Nitin Cheekatla$^c$ (nitincheekatla@gmail.com) \\
				
				\hspace{10pt}
				
				\begin{flushleft}
					\small  
					$^a$ SoCSE, RV University, Bengaluru, India. \\
                    $^b$ School of Engineering and Technology, Vidyashilp University, Bengaluru, India. \\
					$^c$ Dataplex Inc., USA. \\

					\vspace{1cm}
					\textbf{S.H. Shabbeer Basha, Associate Professor, School of Engineering and Technology, Vidyashilp University, Bengaluru, India.} \\
					Email: shabbeer.basha@vidyashilp.edu.in
					
				\end{flushleft}        
			\end{center}
		\end{titlepage}
		
		\title{Are Candidate Models Really needed for Active Learning?}
		
		\author[label1]{Harshini Mridula Mohan}
		\ead{harshinimm.btech22@rvu.edu.in}
		
		\author[label1]{Maanya Manjunath}
		\ead{maanyam.btech22@rvu.edu.in}

        \author[label1]{Vipul Arya}
		\ead{krishnamva.btech22@rvu.edu.in}

        \author[label2]{{S.H.Shabbeer Basha} \corref{cor1}}
		\ead{shabbeer.basha@vidyashilp.edu.in}
        
        \author[label1]{Nitin Cheekatla}
		\ead{nitincheekatla@gmail.com}
		
		\cortext[cor1]{S.H. Shabbeer Basha (Corresponding Author)}
		\address[label1]{SoCSE, RV University, Bengaluru, India.}
        \address[label2]{School of Engineering and Technology, Vidyashilp University, Bengaluru, India.}
		\address[label3]{Dataplex Inc., USA.}
		
		\begin{abstract}
			Deep learning has profoundly impacted domains such as computer vision and natural language processing by uncovering complex patterns in vast datasets. However, the reliance on extensive labeled data poses significant challenges, including resource constraints and annotation errors, particularly in training Convolutional Neural Networks (CNNs) and transformers due to a larger number of parameters. Active learning offers a promising solution to reduce labeling burdens by strategically selecting the most informative samples for annotation. However, the current active learning frameworks are time-intensive; select the samples iteratively with the help of initial candidate models. This study investigates the feasibility of using CNNs and transformers with randomly initialized weights, eliminating the need for initial candidate models while achieving results comparable to active learning frameworks that depend on such candidate models. We evaluate three confidence-based sampling strategies—high confidence (HC), low confidence (LC), and a combination of high confidence in the early stages of training and low confidence at later stages of training (HCLC). Among these, mostly LC demonstrated the best performance in our experiments, showcasing its effectiveness as an active learning strategy without the need for candidate models. Further, extensive experiments verify the robustness of the proposed active learning methods. By challenging traditional frameworks, the proposed work introduces a streamlined approach to active learning, advancing efficiency and flexibility across diverse datasets and domains.
		\end{abstract}
		
		
	\end{frontmatter}
	
	\section{Introduction}
	\label{introduction}
	
Deep learning has revolutionized and impacted numerous fields, from computer vision and natural language processing to healthcare and autonomous systems \cite{sener2018active, vaswani2017attention}. This success is mainly due to their ability to learn and understand complex patterns from vast amounts of data. Convolutional Neural Networks (CNNs) and transformers have been at the forefront of deep learning revolutionaries, mainly in computer vision and natural language processing. However, these neural networks require enormous labeled data for learning the large number of parameters.
	
Deep learning research requires active learning methods which help reduce the adverse effects of training data labeling on model development. The process of annotating extensive datasets poses significant challenges which often result in reduced model performance. Active learning focuses on carefully choosing and labeling the most useful data points. The system operates as a tool that decreases required annotation work while it improves model performance.

Active learning has emerged in the current landscape of deep learning to address the inefficiencies of traditional passive learning methods. This approach
improves model performance and suitability for various domains. The experimental evaluation will help us to assess how well confidence-based sampling methods work while we seek to fix the problems present in standard active learning techniques. The experiments provide essential knowledge that helps us reach our goal of developing active learning techniques which minimize the impact of labels. The different datasets used in our experiments provide opportunities to test how our active learning techniques work in various field applications. Our research will analyze performance through different datasets to find specific patterns and assess how our methods perform in real-world scenarios.

Active learning depends on sample selection because it determines how effectively learning will proceed. Sample selection strategies aim to achieve maximum model improvement through minimal requirements for labeled data. Uncertainty sampling stands as a common method which selects samples that the model exhibits its greatest uncertainty about \cite{lewis1994sequential}, while diversity-based sampling selects multiple data points to avoid selecting identical samples \cite{sener2018active}. Hybrid methods use both uncertainty and diversity elements to choose samples which provide important information while matching the dataset distribution \cite{beluch2018power}. These strategies help to address the trade-off between reducing labeling costs and maintaining model performance especially in fields which require expensive or time-consuming labeling work. The active learning process gains speed through these methods which also enhance model performance when applied to new data.

Traditional active learning frameworks rely on a candidate model (or cold-start model), a model pre-trained on a small initial labeled set ($L_0$) to guide sample selection. This model is used to compute uncertainty, diversity, or other acquisition metrics for active learning on the remaining data. However, this approach introduces computational overhead and potential bias from limited initial training data. Candidate models, by design, aim to make smarter choices for data selection, but this often involves higher processing power to assess model uncertainty, particularly when dealing with complex architectures or ensembles. This additional computation, even in identical training sequences, can make the active learning cycle more resource-intensive with candidate models compared to models initialized with random weights.


\begin{figure}[!tbp]
  \centering
    
    \includegraphics[width=0.75\textwidth]{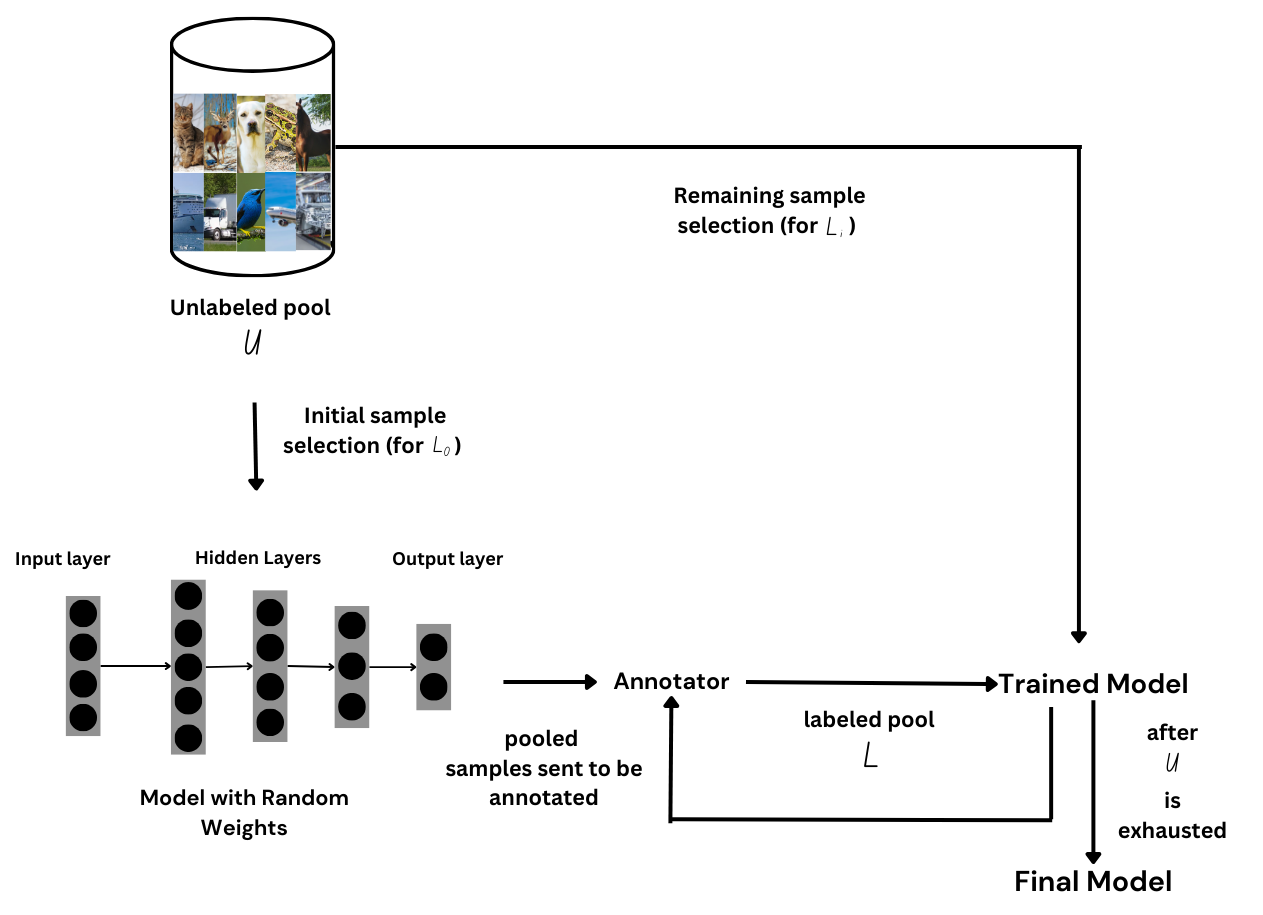}
    \caption{Overview of the proposed Deep Active Learning (DAL) pipeline. Initial samples are selected based on the model confidence scores predicted by the model \( M_0 \) with random weights. The selected samples are annotated and used for training and improving the models iteratively.   }
    \label{fig1}
\end{figure}

In addition to the resource-intensive nature of candidate models, other challenges arise in active learning that can impact its efficacy and generalization ability. One such issue is bias in sample selection, where the focus on uncertain or diverse samples can lead to ignoring certain regions of the feature space, resulting in a model that under-performs on less-represented classes or data distributions \cite{sener2018active}. This is particularly problematic in imbalanced datasets, where certain categories may be systematically ignored during the query process, reducing the model’s overall robustness. Another issue is the cold start problem, where a model trained with very little labeled data may not provide reliable uncertainty estimates early in the active learning process, leading to inefficient query selections. In such cases, it may be necessary to start with a candidate model that is trained on a sufficiently large labeled dataset or employ strategies like hybrid active learning, combining uncertainty sampling with random sampling \cite{huang2010active}. 

Existing active learning methods predominantly rely on pre-trained candidate models to select the most informative samples for annotation, under the assumption that such models provide a beneficial starting point for data selection. However, the Lottery Ticket Hypothesis \cite{frankle2019lottery} challenges this reliance by suggesting that within randomly initialized neural networks there exist sub-networks, or "winning tickets," capable of achieving comparable performance to fully trained models when trained from scratch. This insight provides an alternative viewpoint that enables active learning frameworks to operate efficiently even in the absence of pre-trained candidate models. By directly leveraging the inherent potential of randomly initialized networks, active learning can harness these networks to guide sample selection, bypassing the computational overhead and resource constraints associated with candidate models. This approach not only streamlines the active learning process but also aligns with the principles of efficiency, paving the way for more accessible and flexible frameworks that retain robust performance across diverse datasets.

In summary, this paper’s contributions are as follows:

\begin{itemize}
    \item Thoroughly validated the question, "Are candidate models required for active learning?" 
    \item Developed and tested three uncertainty-based sampling methods for active learning using various deep learning models including a foundation model DinoV2.
    \item Conducted extensive experiments to validate the effectiveness of these methods on model performance.
    \item Provided a comprehensive analysis of the impact of confidence-based sampling on model training, contributing to improved active learning strategies.
\end{itemize}

\section{Related works}
	\label{title_page}

Active learning is a subfield of machine learning (ML), a method developed to achieve significant performance gains training over a few labeled samples. More concretely, active learning aims to select the most informative samples from the unlabeled dataset and provide them for annotation. Active learning is widely adopted to reduce the annotation cost as much as possible while still maintaining the performance \cite{settles2009active, gal2017deep}. In active learning, choosing the right set of samples for annotation and model training is always a challenge. One classic approach is uncertainty sampling; the idea is to select the samples at which model is not very confident \cite{lewis1995sequential}. In this study, it was discovered that training a model with fewer samples can yield results similar to, or even better than, those obtained by training on the entire dataset.

Most of the early works on active learning are based on a survey of \cite{settles2009active}. It covers acquisition functions, namely information-theoretical methods, ensemble approaches \cite{chitta2019training} and uncertainty-based methods \cite{sener2018active}. Speaking about  the advent of big data, the machine learning community has been keen on focusing on big data problems \cite{sundermeyer2012lstm} \cite{kalchbrenner2013recurrent} \cite{sutskever2014sequence}. With increased interest in this domain, new models were developed and existing ones were refined. One such model that gained significant attention is convolutional neural networks (CNN) \cite{krizhevsky2012imagenet}.  CNN was originally developed in 1989 to parse handwritten zip codes \cite{lecun1989backpropagation}, these were adapted and flourished to a point where CNN was able to beat human-level performance, provided sufficient data was given for training \cite{gal2017deep}. However, as CNNs began to excel in tasks requiring large datasets, researchers started to explore methods to improve their efficiency in scenarios where limited labeled data is available. Active learning emerged as a promising approach to address this limitation, leveraging models to identify the most informative samples for annotation. \cite{gal2017deep} used Bayesian CNNs to restore model uncertainty missing in conventional CNNs, which was crucial for active learning. To prove this hypothesis, authors evaluated the performance of active learning using conventional CNNs against Bayesian CNNs and concluded that a proper uncertainty measure enhances active learning  performance.

\cite{sener2018active} introduced the active learning technique \textit{Coreset} which defines active learning problem through Coreset selection. The objective of this method is to select data points which enable model training from this subset to achieve performance equal to that of a model trained on the complete dataset. Their algorithm achieved superior results compared to existing methods for image classification experiments on CIFAR-10, CIFAR-100, and SVHN datasets. They defined \textit{Coreset loss} as the difference between the average empirical loss over the labeled subset and the entire dataset, guiding the selection of new points for labeling to ensure that the model's performance on the labeled subset closely matches its performance on the full dataset. They stated that Bayesian active learning works well with small datasets but has difficulty handling large datasets because of its batch sampling method. The experimental section presents further details about this method's results through a comparison with our developed approaches.

In the last few years, there have been substantial attempts by researchers to extend active learning techniques to deep learning models, which are known as Deep Active Learning (DeepAL) methods \cite{ren2021survey}. A research study carried out by \cite{ren2021survey} presents a comprehensive overview of DeepAL techniques and points out a major drawback of the conventional active learning technique: the inefficiency of the one-by-one sample query strategy \cite{lewis1994sequential,tong2001support}. As this strategy is not scalable for deep learning models, DeepAL adopts a batch-based query strategy, which forms the basis of DeepAL. Based on this, various query optimization techniques have been investigated.

One of the common techniques in Deep Active Learning is the uncertainty-based sampling technique, where the samples are ranked based on the level of uncertainty, and the samples with higher uncertainty are labeled first \cite{nguyen2021uncertainty}. Another technique is the hybrid query method, which considers not only the uncertainty of the samples but also other attributes of the samples \cite{lughofer2012hybrid}.

Another significant area of research in Deep Active Learning (DeepAL) is Deep Bayesian Active Learning (DBAL), which uses Bayesian convolutional neural networks to capture uncertainty and improve query selection \cite{gal2015bayesian}. Some methods focus on density-based techniques, with the goal of finding a core set that can represent the entire data, thus decreasing the cost of annotation \cite{phillips2017coresets}.

For further improvement of DeepAL methods, there have been efforts to automate techniques for active learning query strategy design and deep learning models \cite{ren2021survey}. The applicability of DeepAL techniques, however, depends on the dataset and the technique used. For instance, on the CIFAR-100 dataset with 40\% of the data labeled, the accuracy of the Coreset method \cite{sener2018active} and the variational adversarial active learning method \cite{sinha2019variational} differs by as much as 8\%.

In recent years, there has been a rejuvenation of interest in using active learning for data acquisition for training, improving deep neural networks at scale \cite{citovsky2021batch}. \cite{citovsky2021batch} introduced a batch active learning algorithm called Cluster-Margin. This algorithm can scale the model to very large batch sizes ranging from $10^5$ to $10^6$ while significantly enhancing label efficiency. The core principle of Cluster-Margin is to leverage Hierarchical Agglomerative Clustering (HAC) \cite{citovsky2021batch} to create diverse batches from examples that the model is least confident on. A key benefit of this algorithm is that HAC is executed only once on the entire pool of unlabeled data as an initial step. This setup allows for efficient sampling iterations where the algorithm retrieves clusters from HAC, focusing on the least confident examples. It employs a round-robin scheme to sample across these clusters, ensuring diversity in the selected samples. 

Existing active learning methods rely on a candidate model, which is typically pre-trained on the initial training set. This initial model is used to select the next set of samples for annotation and model training. This approach is used based on the assumption that the model's initial state will be beneficial for selecting the most informative data points for labeling. The Lottery Ticket Hypothesis \cite{frankle2018lottery} presents that subnetworks can perform comparably with the original model when trained from scratch. In this paper, we have examined whether an active learning method can benefit from models with randomly initialized weights, potentially leading to more efficient and scalable active learning frameworks.

\section{Methodology}

\subsection{Preliminaries \& Notations}


The active learning process is defined over a pool of unlabeled data \( U \) and an initially labeled dataset \( L_0 \) which is randomly selected from \( U \). Unlike conventional active learning, our method eliminates the need for a pre-trained candidate model. Instead, we leverage the inherent sample selection capability of randomly initialized networks ($M_0$), reducing bias and computational cost \cite{cao2018review,ramanujan2020s}. The proposed sampling framework leverages \( M_0 \) for the initial iteration and trained models \( M_1, M_2, ..., M_i \) for subsequent $i$ iterations. This process repeats until the annotation budget is exhausted.

The first unlabeled subset \( S_0 \) is selected from \( U \) based on confidence criteria, using either the set of high-confidence samples (\( S_{\text{HC}} \)) or the set of low-confidence samples (\( S_{\text{LC}} \)). High-confidence samples are those for which the model makes high-confidence predictions. In contrast, low-confidence samples are selected from data points where the model shows the greatest uncertainty, offering more informative feedback. Acquisition functions \( \phi_{\text{HC}}(x) \) or \( \phi_{\text{LC}}(x) \) are used to select \( S_0 \), which will be discussed in more detail in the following subsections. Once these initial samples \( S_0 \) are selected, \( M_0 \) is trained on the corresponding labeled set \( L_0 \), producing the first trained model \( M_1 \).

In the \( i^{\text{th}} \) iteration, the model \( M_i \) is trained on the labeled set \( L_i \), which is updated as the new data points \( S_i \) are selected, labeled, and added to the previous iteration training set \( L_{i-1} \). For example, the first iteration \( S_1 \) includes the samples selected from \( U \), leading to an updated labeled set \(L_1 = L_0 \cup \hspace{0.1cm} labeled(S_1)\). Here, \( \text{labeled}(S_1) \) indicates the labeled data of \( S_1 \). In the second iteration, additional samples \( S_2 \) are selected, forming \(L_2 = L_1 \cup \hspace{0.1cm} labeled(S_2)\). This process continues iteratively, where the labeled set is updated at each step as \(L_i = L_{i-1} \cup \hspace{0.1cm} labeled(S_i)\). The acquisition functions \( \phi_{\text{HC}}(x) \) and \( \phi_{\text{LC}}(x) \) are applied to the remaining unlabeled data \( U \) in each iteration, identifying the next set of high or low confidence samples. High-confidence samples reinforce the model to understand familiar patterns, while low-confidence samples help the model learn from areas of greater uncertainty. As the process iterates, the labeled set \( L \) grows, and new samples are annotated, enriching the dataset. This iterative process allows the model to evolve from its initial state \( M_0 \) into progressively refined versions \( M_i \), systematically improving performance by exploring both high-confidence and low-confidence samples within the unlabeled data. Figure ~\ref{fig1} visually illustrates this process, where the initial model \( M_0 \) with random weights begins training on an initial sample set. The selected samples are annotated by an annotator and added to the labeled pool \( L \), forming the training set for subsequent models. The process iterates until all unlabeled samples are exhausted or the annotation budget is exhausted. High-confidence samples help reinforce familiar patterns, while low-confidence samples enable the model to learn from uncertain areas, systematically improving its performance.

\begin{algorithm}
\caption{Proposed Active Learning Pipeline}
\begin{algorithmic}
\State // Input: Unlabeled pool \( U \), model with random weights \( M_0 \), annotation budget B
\State Obtain model confidence scores for all samples \( \in  U \) \\

 // Use \( \phi_{\text{LC}}(x) \) to select uncertain samples or \( \phi_{\text{HC}}(x) \) to select easy initial samples based on the active learning sampling method
\State Compute \( \phi_{\text{LC}}(x) \)  or \( \phi_{\text{HC}}(x) \) for \( x \in U \)
\State // Select initial samples based on the active learning sampling method to train model \( M_0 \) on \(L_0\)
\State \( S_0 \gets \text{Select}(U, \phi) \)
\State Set \( L_0 = labeled(S_0) \)
\State // Update unlabeled pool
    \State \( U \gets U - S_0 \)
\State \( M_0 \gets \text{Train}(L_0) \)
\State \( i \gets 1 \)

\While {B is not exhausted and \( U \neq \emptyset \)}

    \State Compute model confidence scores for remaining samples for \( x \in U - S_{i-1} \)
    \State  // Use \( \phi_{\text{LC}}(x) \) to select uncertain samples or \( \phi_{\text{HC}}(x) \) to select easy samples based on the active learning sampling method
    \State Compute \( \phi_{\text{LC}}(x) \) or  \( \phi_{\text{HC}}(x) \) for \( x \in U - S_{i-1} \)
    
    \State // Select a batch of samples based on acquisition functions
    \State \( S_i \gets \text{Select}(U, \phi) \)
    
    \State // Update labeled set
    \State \( L_i \gets L_{i-1} \cup \text{labeled}(S_i) \)
    
    \State // Update unlabeled pool
    \State \( U \gets U - S_i \)

    \State // Train model on the labeled set
    \State \( M_i \gets \text{Train}(L_i) \)
\EndWhile

\State // Output: Final model trained on labeled set
\State \textbf{Output:} \( M_i \)
\end{algorithmic}
\label{alg1}
\end{algorithm}

\subsection{Key contributions}

This work investigates whether pre-trained models trained using the initial dataset often termed candidate models are essential for deep active learning. Typically, active learning methods rely on candidate models to give the learning process a head start, using knowledge gained on the initial training set to improve sample selection \cite{huang2010active}. However, this study demonstrates that starting an active learning process with random weighted models can achieve competitive results without the need for such models. This study also helps the researchers to explore more on the use of sub-networks for efficient deep active learning.

Excluding the candidate models not only makes the active learning process efficient but also avoids bias in the learning process. These candidate models are often based on small samples that may not match the specific task at hand, potentially leading to poor choices in data selection. By using random initialization, the model is forced to learn directly from the labeled data. This also ensures that the representations learned by the model are more focused and specific to the task being solved.

Additionally, this work shows that effective sampling can be performed without relying on candidate models. In many active learning methods, sampling strategies (such as selecting uncertain or diverse samples) work best when the model is already well calibrated, which is typically achieved through training on the initial training set. However, this research demonstrates that even models with random weights can choose the right samples, meaning that strong initial knowledge from candidate models is not always necessary for deep active learning.

In summary, the proposed method eliminates the need for a trained candidate model in active learning, which is a key distinction from traditional methods (e.g., uncertainty sampling \cite{lewis1995sequential}, query-by-committee \cite{seung1992query}, or Coreset \cite{sener2018active} approaches). The novelty lies in leveraging the inherent capability of randomly initialized dense networks to identify informative samples, supported by the following observations:

\begin{itemize}
    \item Model-Free Sample Selection: Unlike traditional methods that require a well-formed candidate model to estimate uncertainty or diversity, the proposed approach relies on the insight that larger neural networks are able to rank samples effectively even with random initialization \cite{cao2018review,ramanujan2020s}.
    \item The traditional method is vulnerable to bias caused by the training of the candidate model. The proposed approach overcomes this issue by not depending on any pre-trained model, which improves robustness in the early stages of active learning.
    \item Scalability: In larger networks, there is a naturally strong signal-to-noise ratio in the gradients or activations, which makes sample selection easier without training. This is in line with the Lottery Ticket Hypothesis, which states that randomly initialized networks have good structural priors for sample selection, thus reducing the need for candidate models \cite{frankle2018lottery}.
\end{itemize}

\subsection{Evaluated sampling methods}

This paper evaluates the impact of three different sampling approaches on model performance. The High Confidence (HC) approach focuses on the selection of points for which the model has the highest confidence, while the Least Confidence (LC) approach focuses on points for which the model has the lowest confidence. Finally, the hybrid approach, High Confidence and Least Confidence (HCLC), combines both approaches by selecting points from both ends of the confidence spectrum. The next sections will provide a more in-depth look into these approaches.

In Algorithm \ref{alg1}, the acquisition function is chosen based on the desired sampling strategy. Low-Confidence (LC) sampling $\phi_{\text{LC}}(x) = 1 - \max_k P(y=k \mid x)$ is designed for uncertainty-based exploration, selecting samples where the model is least confident. Generally, this strategy is most effective in later training stages when the model can benefit from refining boundaries on challenging examples. High-Confidence (HC) sampling $\phi_{\text{HC}}(x) = \max_k P(y=k \mid x)$ selects easy, reliable samples for stability-based exploitation, particularly valuable in the initial cycle when the model has random weights and needs a robust foundation. The hybrid HCLC strategy optimally combines both. This phased approach ensures the model first learns from confident patterns before tackling ambiguous data, improving both convergence speed and final accuracy. In practice, the choice depends on the application—HC sampling is preferred in class-imbalanced scenarios to ensure minority class representation, while LC sampling works well in balanced datasets for maximum information gain. We provide more detailed discussion on these sampling methods in sections \ref{HC_section}, \ref{LC_section}, and \ref{HCLC_section}.

\subsubsection{High Confidence (HC) Sampling}
\label{HC_section}

The high-confidence sampling method begins with the selection of 10,000 samples from the pool of unlabeled samples based on the confidence scores of the model. The selected samples are those for which the model has the highest level of confidence after the initial training, based on the high-confidence (HC) criterion:

\[
\boldsymbol{\phi}_{\text{HC}}(x) = \max_{k} P(y = k \mid x)
\]

This acquisition function represents choosing samples with maximum predicted probability score. Following the initial round, an additional $5,000$ high-confidence samples are incrementally added during each subsequent iteration. 

\subsubsection{Low Confidence (LC) Sampling}
\label{LC_section}
In contrast to HC sampling, the LC sampling method prioritizes choosing samples where the model shows the least confidence in its predictions. Following is the Low Confidence (LC) criterion:

\[
\boldsymbol{\phi}_{\text{LC}}(x) = 1 - \max_{k} P(y = k \mid x)
\]

Here, \( 1-P(y = k \mid x) \) is the least predicted probability for class \( k \), capturing the model's uncertainty. Similar to HC, this approach starts with an initial subset of $10,000$ low-confidence samples, and 5,000 additional low-confidence samples are added repeatedly after each iteration.

\subsubsection{High Confidence Least Confidence (HCLC) Sampling}
\label{HCLC_section}
The HCLC approach, as its name suggests, combines elements of both high-confidence (HC) and low-confidence (LC) approaches. The process begins with an initial selection of 10,000 high-confidence samples based on the HC measure to train the model on data it is already familiar with. In the subsequent rounds, low-confidence samples are added based on the LC measure to provide the model with exposure to more difficult and uncertain samples. By gradually adding these low-confidence samples, the model improves its ability to generalize and make correct predictions for uncertain data points. This complementary strategy leverages the benefits of both HC and LC approaches—beginning with reliable data to build a strong foundation and then adding uncertain samples to improve robustness. As such, the HCLC approach is effective in optimizing the performance of the model by maintaining a well-calibrated training process.

\section{Experimental Results} \label{Proposed Method}
This section describes the experimental design used to evaluate the efficacy of the proposed active learning approaches. The proposed sampling strategies include LC (Low Confidence), HC (High Confidence), and HCLC (High Confidence + Low Confidence). The experiments are carried out on various benchmark datasets such as CIFAR-10, CIFAR-100, SVHN, and TinyImageNet using different deep neural network models for image classification. PASCAL VOC has been utilized for object detection task. The aim of the experiments is to compare the performance of the proposed approaches with other existing active learning methods.

As shown in Table \ref{tab:CIFAR10}, we comprehensively evaluate our proposed active learning methods against a wide range of existing approaches on CIFAR-10 across seven different architectures. \textbf{DenseNet-121:} Our HCLC method achieves 93.08\% accuracy, outperforming all compared baselines. This includes substantial improvements over Gaussian Switch Sampling (72.00\%) \cite{benkert2023gaussian}, Triplet AL (76.00\%) \cite{sundriyal2021semi}, Uncertainty Estimation (91.00\%) \cite{nguyen2021uncertainty}, ALFA-Mix (67.00\%) \cite{parvaneh2022active}, and TLC (85.96\%) \cite{liu2023imbalanced}. We integrated results for \cite{rakesh2021efficacy}, \cite{beluch2018power}, and COPS \cite{lin2023optimal}, which achieved accuracies of $82.00\%$, $91.20\%$, and $80.50\%$ respectively. Notably, our approach also surpasses Band-limited Training (92.00\%) \cite{dziedzic2019band}, a method that similarly does not require a candidate model, demonstrating the effectiveness of our confidence-based sampling strategies even within the candidate-model-free paradigm.  To ensure a robust and standardized benchmark across diverse neural architectures, we have reproduced the active learning results of BADGE \cite{ash2019deep}, BAIT \cite{ash2021gone}, and GLISTER specifically for the DenseNet-121. Our reproduced results on CIFAR-10 indicate that while GLISTER ($92.86\% \pm 0.12\%$) and BADGE ($91.24\% \pm 0.09\%$) maintain high performance, BAIT ($70.91\% \pm 0.20\%$) appears more sensitive to the DenseNet-121. 
\textbf{ResNet-56:} Both LC and HCLC methods achieve 91.60\% accuracy, outperforming Superconvergence (91.10\%) \cite{smith2019super} and Adversarial Sampling (89.00\%) \cite{sinha2019variational}. The Double-Cross Attack method achieves 78.44\% \cite{vicarte2021double}, significantly lower than our approaches, highlighting the robustness of confidence-based sampling. The comparative scope was broadened by adding PrAC \cite{zhang2021efficient} ($92.00\%$) and the recent PruneFuse \cite{kousar2025pruning}, which demonstrates a high accuracy of $93.65\%$.
\textbf{VGG-16:} Our LC method achieves 94.21\% accuracy, the highest among all compared methods on this architecture. This represents improvements of 3.21\% over FF-Active (91.00\%) \cite{geifman2017deep}, 4.05\% over VAAL (90.16\%) \cite{sinha2019vaal}, and 4.21\% over both Coreset variants (90.00\%) \cite{sener2018active}. The comparison with BADGE (82.00\%) \cite{ash2019deep} and \cite{jung2023simple} (81.44\%) further demonstrates the superiority of our approach, with improvements exceeding 12\%. Even at reduced budgets (20K, 40K samples), our methods show competitive performance (84.26\% and 91.89\% respectively). \textbf{ResNet-18:} Our LC method achieves 93.53\% accuracy, outperforming DEAL (88.00\%) \cite{hemmer2022deal}, VAAL (81.00\%) \cite{sinha2019vaal}, ST-CoNAL (5K) (83.05\%) \cite{baik2022st}, and SSAL (92.7\%) \cite{li2022empirical}. At the 40K budget setting, our method achieves 93.69\%, significantly exceeding BADGE (59.00\%) \cite{ash2019deep} and demonstrating consistent improvement across budget levels. Ensemble AL achieves 96.30\%, the highest reported accuracy on this architecture, though it requires a candidate model and substantial computational overhead. \textbf{Swin Transformer:} Our LC method achieves 86.23\% accuracy, substantially outperforming MESA (81.30\%) \cite{pan2021mesa} and NHT (78.00\%) \cite{zhang2022nested}. The 4.93\% improvement over MESA demonstrates that candidate-model-free sampling is effective even on modern transformer architectures. \textbf{ViT-Small:} Our methods achieve 82.70-83.92\% accuracy, competitive with Knowledge Distillation (73.80\%) \cite{zhang2022knowledge}, Bootstrapping ViTs (87.32\%) \cite{zhang2022bootstrapping}, and Interpretability-Aware ViT (88.93\%) \cite{yu2023x}. While the pre-trained transformer baselines achieve higher accuracy, they require extensive pre-training—our approach achieves strong performance with zero pre-training overhead. \textbf{MobileNetV2:} Our LC and HCLC methods both achieve 94.16\% accuracy, dramatically outperforming FPGA (86.00\%) \cite{bouguezzi2021efficient}, AOT (80.00\%) \cite{jindal2024army}, and PA-CL (82.00\%) \cite{chen2024manipulating}. The 8-14\% improvement on this lightweight architecture is particularly noteworthy, as it demonstrates that our approach is especially effective when computational resources are limited. To demonstrate the generalizability of our "model-free" approach on lightweight architectures, we added a comparison with COPS (Lin et al., 2023), which reported $74.00\%$ accuracy.

\begin{figure}[t]
  \centering
  \begin{subfigure}[b]{0.5\textwidth}
    \centering
    \includegraphics[width=\textwidth]{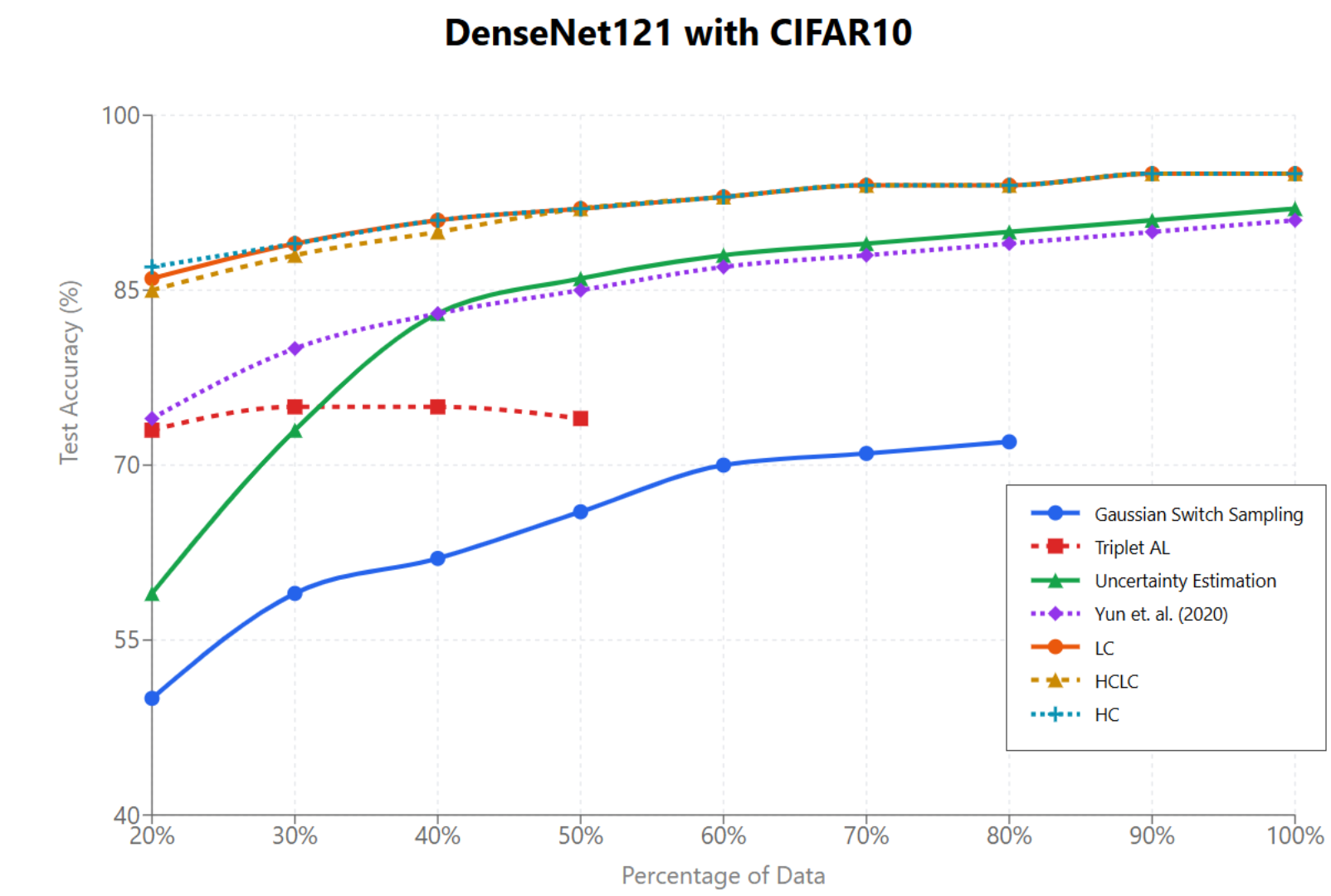}
  \end{subfigure}
  \hfill
  \begin{subfigure}[b]{0.48\textwidth}
    \centering
    \includegraphics[width=\textwidth]{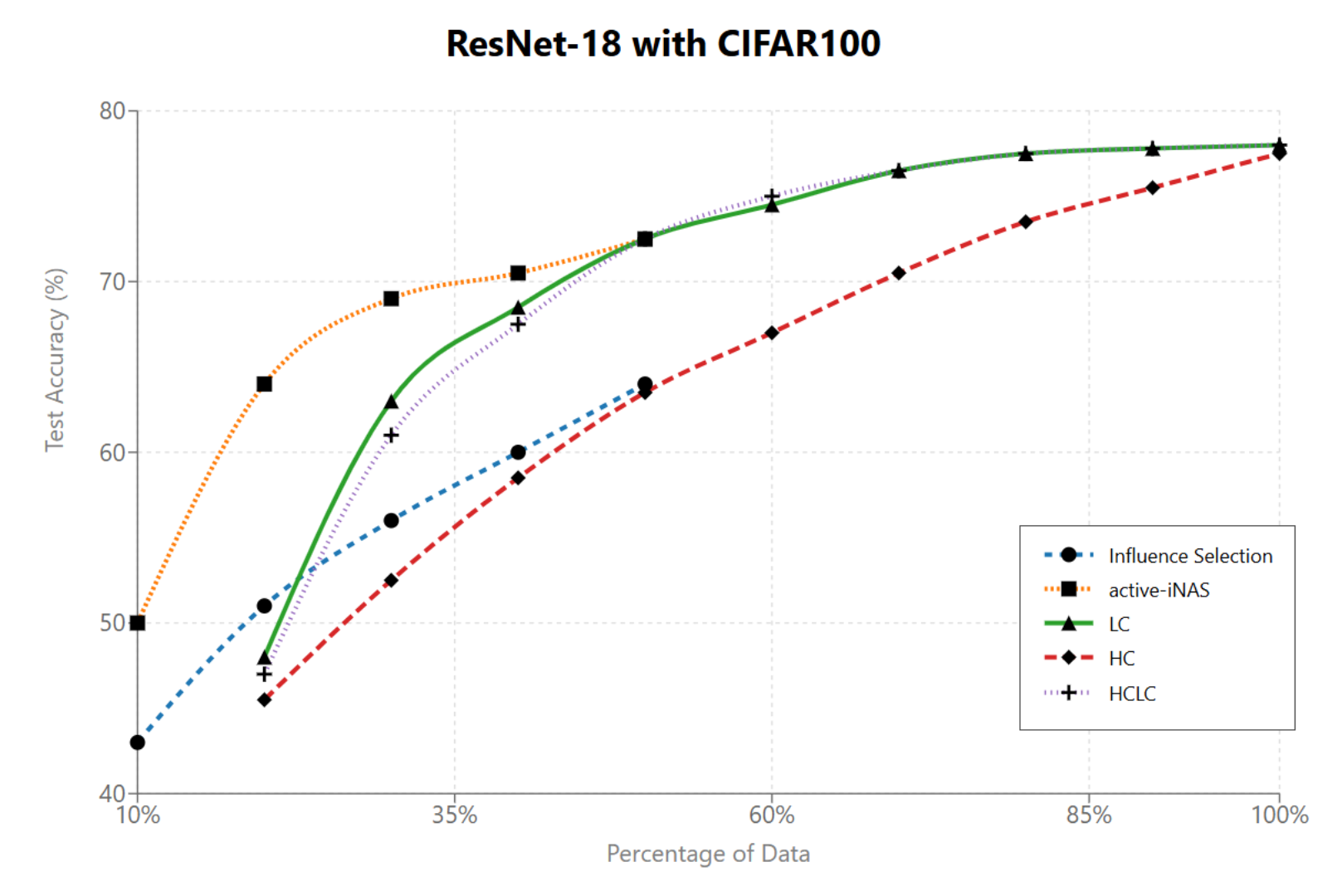}
  \end{subfigure}
  \caption{Comparison of model performances at different proportions of training data. CIFAR-10 and CIFAR-100 datasets. Left: Test accuracy on CIFAR-10 using DenseNet-121. Right: Test accuracy on CIFAR-100 using ResNet-18.}
  \label{fig:cifar-comparison}
\end{figure}

As shown in Figure \ref{fig:cifar-comparison}, the DenseNet-121 model achieved an accuracy of 92.87\% using the LC approach, and by combining the approaches, the accuracy was improved to 93.08\%, performing better than other approaches like Uncertainty Estimation (91\%) \cite{nguyen2021uncertainty} and Triplet AL (76\%). 

Table \ref{tab:CIFAR100_candidate_time} presents the comparison of active learning methods on the more challenging CIFAR-100 dataset across five architectures. The increased class complexity provides a rigorous test of our candidate-model-free sampling strategies. \textbf{DenseNet-121: } Our LC method achieves 71.65\% accuracy, substantially outperforming Bayesian Neural Networks (60.00\%) \cite{rakesh2021efficacy} and TLC (62.05\%) \cite{liu2023imbalanced} by margins of 9-11\%. Even at the 10K budget setting, our method achieves 59.03\%, exceeding \cite{yun2020weight} (57.43\%) and \cite{rakesh2021efficacy} ($59.00\%$) while requiring no additional time for training candidate model. These results demonstrate that our approach scales effectively with additional annotations on complex datasets. On the CIFAR-100 dataset, the reproduced BADGE baseline achieves $71.82\% \pm 0.20\%$, serving as a rigorous benchmark for evaluating our proposed candidate-model-free approach. \textbf{ResNet-56:} Our methods achieve 66.22-66.39\%, competitive with but slightly below Passive Batch Injection Training (71.60\%) \cite{singh2020passive} and Linear Feature Disentanglement (70.73\%) \cite{he2022exploring}. This suggests that on certain architectures, candidate-model-free methods face limitations. However, our methods still substantially outperform Superconvergence (60.80\%) \cite{smith2019super} and Adversarial Sampling (57.50\%) \cite{sinha2019variational}. We expanded the comparison to include \cite{kousar2025pruning} ($67.87\%$) and PrAC \cite{zhang2021efficient} ($71.20\%$), ensuring that our results are evaluated against the latest literature. \textbf{VGG-16:} Our LC method achieves 66.46\% accuracy, outperforming FF-Active (66.00\%) \cite{geifman2017deep} and dramatically exceeding VAAL (47.60\%) \cite{sinha2019vaal} by +18.86\%. The comparison with Coreset-WSM (70.00\%) \cite{sener2018active} shows our approach is competitive while requiring no additional time for training candidate model. \textbf{ResNet-18:} Our HCLC method achieves 73.42\% accuracy, the highest among all compared methods except Ensemble AL (81.86\%) \cite{chitta2019training}. This represents improvements of 2.92\% over active-iNAS (70.50\%) \cite{geifman2019deep} and 9.42\% over Influence Selection (64.00\%) \cite{liu2021influence}. At the 10K budget, our method achieves 59.01\%, outperforming ST-CoNAL \cite{baik2022st} (57.49\%), CoreGCN \cite{caramalau2021sequential} (56.5\%), UncertainGCN \cite{caramalau2021sequential} (55\%) and comparable results compared to \cite{yun2020weight} (59.73\%). \textbf{MobileNetV2:} This architecture shows the most dramatic improvements. Both our LC and HCLC methods achieve 73.79\% accuracy, outperforming all candidate-model baselines by substantial margins: 7.52\% over Coreset (66.27\%) \cite{sener2018active}, +9.01\% over Glister (64.78\%) \cite{killamsetty2021glister}, and +9.34\% over Moderate Coreset (64.45\%) \cite{xia2022moderate}. The 9\%+ improvement on this lightweight architecture demonstrates that our approach is particularly effective when model capacity is limited.



\begin{table}[!ht]
\centering
\caption{Comparison of deep active learning methods on CIFAR-10.}
\label{tab:CIFAR10_candidate_time}

\centering
\scriptsize
\resizebox{\linewidth}{!}{
\begin{tabular}{|l|p{60mm}|c|c|c|c|}
\hline
\textbf{Model} & \textbf{Method} & \textbf{Candidate Model Used?} & \textbf{Time Saved (hrs)} & \textbf{Annotation Simulation Time (hrs)} & \textbf{Accuracy (\%)} \\ \hline

DenseNet-121 
& Gaussian Switch Sampling \cite{benkert2023gaussian} & Yes & 0.00 & - & 72.00 \\
& Triplet AL \cite{sundriyal2021semi} & Yes & 0.00 & - & 76.00 \\
& Uncertainty Estimation \cite{nguyen2021uncertainty} & Yes & 0.00 & - & 91.00 \\

& Transfer Learning Classifier (TLC) \cite{liu2023imbalanced}  & Yes & 0.00 & - & 85.96 \\
& \cite{yun2020weight}  & Yes & 0.00 & - & 91.70 \\

& \cite{rakesh2021efficacy} (8K) & Yes & 0.00 & - & 82.00 \\

& Band-limited Training \cite{dziedzic2019band} & No & - & - & 92.00 \\

&  Glister \cite{killamsetty2021glister} & Yes & 0.00 & 9.49 & $92.86 \pm 0.12$ \\

&  BADGE \cite{ash2019deep} & Yes & 0.00 & 7.60 & $91.24 \pm 0.09$ \\

&  BAIT \cite{ash2021gone} & Yes & 0.00 & 9.22 & $70.91 \pm 0.20$ \\

&  \cite{beluch2018power} (14.5K) & Yes & 0.00 & - & 91.20 \\

&  COPS \cite{lin2023optimal} (10K) & Yes  & 0.00 & - & 80.50 \\

& Ours (LC) (10K) & No & 0.16 & 0.90 & 91.37 $\pm$ 0.20 \\

& Ours (LC)  & No & 0.16 & 5.09 & 92.87 $\pm$ 0.07 \\

& Ours (HC) & No & 0.16 & 5.09 & 92.72 $\pm$ 0.16 \\
& \textbf{Ours (HCLC)} & \textbf{No} & \textbf{0.16} & \textbf{5.09} & \textbf{93.08 $\pm$ 0.23} \\ \hline

ResNet-56 
& Double-Cross Attacks \cite{vicarte2021double} & Yes & 0.00 & - & 78.44 \\
& Superconvergence \cite{smith2019super} & No & - & - & 91.10 \\
& Adversarial Sampling \cite{sinha2019variational} & Yes & 0.00 & - & 89.00 \\

&  PrAC \cite{zhang2021efficient} & Yes  & 0.00 & - & 92.00 \\

&  \textbf{PruneFuse \cite{kousar2025pruning}} & \textbf{Yes}  & \textbf{0.00} & \textbf{-} & \textbf{93.65} \\

&  COPS \cite{lin2023optimal} (10K) & Yes  & 0.00 & - & 74.50 \\

&  Ours (LC)(10K) & No  & 0.26 & 1.74 & 79.76 \\

& Ours (LC) & No & 0.26 & 8.33 & 91.60 $\pm$ 0.12 \\
& Ours (HC) & No & 0.26 & 8.33 & 91.30 $\pm$ 0.09 \\
& Ours (HCLC) & No & 0.26 & 8.33 & 91.60 $\pm$ 0.12 \\ \hline

VGG-16 
& FF-Active \cite{geifman2017deep} & Yes & 0.00 & -  & 91.00 \\
& VAAL \cite{sinha2019vaal} & Yes & 0.00 & - & 90.16 \\
& Coreset -- WSM \cite{sener2018active} & Yes & 0.00 & - & 90.00 \\
& Coreset -- FSM \cite{sener2018active} & Yes & 0.00 & - & 90.00 \\
& BADGE  \cite{ash2019deep} (40K) & Yes & 0.00 & - & 82.00 \\
& \cite{jung2023simple} (25K) & Yes & 0.00 & - & 81.44 \\
& Discriminative Active Learning (DAL)  \cite{gissin2019discriminative} (25K) & Yes & 0.00 & - & 84.00 \\
& Ours (LC) (20K) & No & 0.5 & 3.60 & 84.26 \\
& Ours (LC) (40K) & No & 0.5 &  6.25 & 91.89 \\
& \textbf{Ours (LC)} & \textbf{No} & \textbf{0.5} & \textbf{8.77} & \textbf{ 94.21 $\pm$ 0.14 } \\
& Ours (HC) & No & 0.5 & 8.77 & 93.94 $\pm$ 0.19 \\
& Ours (HCLC) & No & 0.5 & 8.77 & 94.20 $\pm$ 0.11 \\ \hline

ResNet-18 
& DEAL \cite{hemmer2022deal} & Yes & 0.00 & - & 88.00 \\
& VAAL \cite{sinha2019vaal} & Yes & 0.00 & - & 81.00 \\
& ST-CoNAL  \cite{baik2022st} (5K) & Yes & 0.00 & - & 83.05 \\
& LC (Ours) (5K) & No & 0.25 & 0.25 & 81.27 $\pm$ 0.12 \\
& \cite{yun2020weight} (10K)  & Yes & 0.00 & - & 90.43 \\

& ActiveLossNet \cite{xie2023effective} (10K)  & Yes & 0.00 & - & 77.22 \\

& CoreGCN \cite{caramalau2021sequential} (10K) & Yes & 0.00 & - & 91.50 \\

& UncertainGCN \cite{caramalau2021sequential} (10K) & Yes & 0.00 & - & 91.00 \\

& Ours (LC) (10K)  & No & 0.25 & 0.6 & 90.12 $\pm$ 0.07 \\

& BADGE (40K) \cite{ash2019deep} & Yes & 0.00 & - & 59.00 \\
& Ours (LC) (40K) & No & 0.25 & 3.40 & $92.69 \pm 0.09$ \\
& Ours (HC) (40K) & No & 0.00 & 3.40 & $92.69 \pm 0.11$ \\
& \textbf{Ensemble AL \cite{chitta2019training}} & \textbf{Yes} & \textbf{0.00} & \textbf{-} & \textbf{96.18} \\
& Gaussian Switch Sampling \cite{benkert2023gaussian} & Yes & 0.00 &-  & 66.11 \\
& SSAL \cite{li2022empirical} & Yes & 0.00 & - & 92.7 \\
& Ours (LC) & No & 0.25 & 4.27 & 93.53 $\pm$ 0.08 \\
& Ours (HC) & No & 0.25 & 4.27 & 93.28 $\pm$ 0.03 \\
& Ours (HCLC) & No & 0.25 & 4.27 & 93.48 $\pm$ 0.12 \\ \hline

SWIN 
& MESA \cite{pan2021mesa} & Yes & 0.00 & - & 81.30 \\
& NHT \cite{zhang2022nested}  & No & - & - & 78.00 \\
& \textbf{Ours (LC)} & \textbf{No} & \textbf{0.06} & \textbf{3.82} & \textbf{ 86.23 $\pm$ 0.10} \\
& Ours (HC) & No & 0.06 & 3.82 & 83.88 $\pm$ 0.35 \\
& Ours (HCLC) & No & 0.06 & 3.82 & 85.80 $\pm$ 0.02 \\ \hline

ViT-Small 
& Knowledge Distillation \cite{zhang2022knowledge} & Yes & 0.00 & - & 73.80 \\
& Bootstrapping ViTs \cite{zhang2022bootstrapping} & Yes & 0.00 & - & 87.32 \\
& \textbf{Interpretability-Aware ViT \cite{yu2023x}} & \textbf{Yes} & \textbf{0.00} & \textbf{-} & \textbf{88.93} \\
& Ours (LC) & No & 0.52 & 3.67 & 83.92 $\pm$ 0.36 \\
& Ours (HC) & No & 0.52 & 3.67 & 82.70 $\pm$ 0.18 \\
& Ours (HCLC) & No & 0.52 & 3.67 & 83.81 $\pm$ 0.21 \\ \hline

MobileNetV2 
& FPGA \cite{bouguezzi2021efficient} & Yes & 0.00 & - & 86.00 \\
& AOT \cite{jindal2024army} & Yes  & 0.00 & - & 80.00 \\
& PA-CL \cite{chen2024manipulating} & Yes & 0.00 & - & 82.00 \\

&  COPS \cite{lin2023optimal} (10K) & Yes  & 0.00 & - & 74.00 \\

&  Ours (LC)(10K) & No  & 0.52 & 1.50 & 82.53 \\

& \textbf{Ours (LC)} & \textbf{No} & \textbf{0.20} & \textbf{6.52} & \textbf{94.16 $\pm$ 0.03} \\
& Ours (HC) & No & 0.20 & 6.52& 92.84 $\pm$ 0.23 \\
& \textbf{Ours (HCLC)} & \textbf{No} & \textbf{0.20} & \textbf{6.52} & \textbf{ 94.16 $\pm$ 0.03} \\ \hline

CNN (6-layer) 
& LAL-IGradV-VAE \cite{flesca2024meta} & Yes & 0.00 & - & 83.10 \\ \hline

\end{tabular}
}
\begin{minipage}{\linewidth}
\footnotesize 
\tiny \textit{Note:} DenseNet-121 results for BADGE, BAIT, and GLISTER (highlighted in blue) are reproduced using official code-bases under the same experimental conditions (CIFAR-10, DenseNet-121, 3 runs). All other baseline results are taken from their respective original publications.
\end{minipage}
\label{tab:CIFAR10}
\end{table}

\begin{table}[!ht]
\centering
\caption{Comparison of deep active learning methods on CIFAR-100}
\label{tab:CIFAR100_candidate_time}
\footnotesize
\resizebox{\textwidth}{!}{
\begin{tabular}{|c|p{77mm}|c|c|c|c|}
\hline
\textbf{Model} & \textbf{Method} & \textbf{Candidate Model Used?} & \textbf{Time Saved (hrs)} & \textbf{Annotation Simulation Time (hrs)} & \textbf{Accuracy (\%)} \\ \hline

\multirow{5}{*}{DenseNet-121} 
& Bayesian Neural Networks \cite{rakesh2021efficacy} & Yes & 0.00 & - & 60.00 \\

&  Glister \cite{killamsetty2021glister} & Yes & 0.00 & 4.96 & $68.36 \pm 0.12$ \\

&  \textbf{BADGE \cite{ash2019deep}} & \textbf{Yes} & \textbf{0.00} & \textbf{7.90} & \textbf{$71.82 \pm 0.20$} \\

&  BAIT \cite{ash2021gone} & Yes & 0.00 & 9.75 & $65.91 \pm 0.19$ \\

& \cite{rakesh2021efficacy} (10K)  & Yes & - & - & 59.00 \\

& TLC \cite{liu2023imbalanced}  & Yes & 0.00 & - & 62.05 \\
& Ours (LC) (10K)  & No & 0.19 & 0.4 & 59.03 \\

& \cite{yun2020weight} (10K)  & Yes & 0.00 & - & 57.43 \\
& Ours (LC) & No & 0.19 & 5.27 & 71.65 $\pm$ 0.15 \\
& Ours (HC) & No & 0.19 & 5.27 & 70.98 $\pm$ 0.19 \\
& Ours (HCLC) & No & 0.19 & 5.27 & 71.33 $\pm$ 0.24 \\ \hline

\multirow{7}{*}{ResNet-56} 
& Superconvergence \cite{smith2019super} & No & - & - & 60.80 \\
& Adversarial Sampling \cite{sinha2019variational} & Yes & 0.00 & - & 57.50 \\
& \textbf{Passive Batch Injection Training \cite{singh2020passive}} & \textbf{No} & \textbf{-} & \textbf{-} & \textbf{71.60} \\

&  PrAC \cite{zhang2021efficient} & Yes  & 0.00 & - & 71.20 \\

&  PruneFuse \cite{kousar2025pruning} & Yes  & 0.00 & - & 67.87 \\

& Linear Feature Disentanglement \cite{he2022exploring} & No & - & - & 70.73 \\
& Ours (LC) & No & 0.31 & 8.51 & 66.22 $\pm$ 0.21 \\
& Ours (HC) & No & 0.31 & 8.51 & 66.30 $\pm$ 0.47 \\
& Ours (HCLC) & No & 0.31 & 8.51 & 66.39 $\pm$ 0.06 \\ \hline

\multirow{8}{*}{VGG-16} 
& FF-Active \cite{geifman2017deep} & Yes & 0.00 & - & 66.00 \\
& VAAL \cite{sinha2019vaal} & Yes & 0.00 & - & 47.60 \\
& \textbf{Coreset -- WSM} \cite{sener2018active} & \textbf{Yes} & \textbf{0.00} & \textbf{-} & \textbf{70.00} \\
& Coreset -- FSM \cite{sener2018active} & Yes & 0.00 & - & 65.00 \\
& FF-Active (variant) \cite{geifman2017deep} & Yes & 0.00 & - & 67.00 \\
& Ours (LC) & No & 0.57 & 9.10 & 66.46 $\pm$ 0.63 \\
& Ours (HC) & No & 0.57 & 9.10 & 64.87 $\pm$ 0.35 \\
& Ours (HCLC) & No & 0.57 & 9.10 & 66.11 $\pm$ 0.16 \\ \hline

\multirow{6}{*}{ResNet-18} 
& Influence Selection \cite{liu2021influence} & Yes & 0.00  & - & 64.00 \\
& active-iNAS \cite{geifman2019deep} & Yes & 0.00 & - & 70.50 \\
& ST-CoNAL \cite{baik2022st} (10K) & Yes & 0.00 & - & 57.49 \\
& CoreGCN \cite{caramalau2021sequential} (10K) & Yes & 0.00 & - & 56.50 \\

& UncertainGCN \cite{caramalau2021sequential} (10K) & Yes & 0.00 & - & 55.00 \\

& \cite{yun2020weight} (10K)  & Yes & 0.00 & - & 59.73 \\
& Ours (LC) (10K) & No & 0.10 & 3.88 & 59.01 $\pm$ 0.22 \\
& \textbf{Ensemble AL \cite{chitta2019training}} & \textbf{Yes} & \textbf{0.00} & \textbf{-} & \textbf{81.86} \\ 

& Ours (LC) & No & 0.10 & 3.88 & 73.24 $\pm$ 0.19 \\
& Ours (HC) & No & 0.10 & 3.88 & 71.97 $\pm$ 0.05 \\
& Ours (HCLC) & No & 0.10 & 3.88 & 73.42 $\pm$ 0.24 \\ \hline

\multirow{7}{*}{MobileNetV2} 
& LAL-IGradV-VAE \cite{flesca2024meta} & Yes & 0.00 & - & 64.50 \\
& Glister \cite{killamsetty2021glister} & Yes & 0.00 & - & 64.78 \\
& Coreset \cite{sener2018active} & Yes & 0.00 & - & 66.27 \\
& Moderate Coreset \cite{xia2022moderate} & Yes & 0.00 & - & 64.45 \\
& \textbf{Ours (LC)} & \textbf{No} & \textbf{0.25} & \textbf{6.63} & \textbf{73.79 $\pm$ 0.13} \\
& Ours (HC) & No & 0.25 & 6.63 & 72.82 $\pm$ 0.13 \\
& \textbf{Ours (HCLC)} & \textbf{No} & \textbf{0.25} & \textbf{6.63} & \textbf{73.79 $\pm$ 0.13} \\ \hline

\end{tabular}
}
\begin{minipage}{\linewidth}
\footnotesize 
\tiny \textit{Note:} DenseNet-121 results for BADGE, BAIT, and GLISTER (highlighted in blue) are reproduced using official code-bases under the same experimental conditions (CIFAR-100, DenseNet-121, 3 runs). All other baseline results are taken from their respective original publications.
\end{minipage}
\label{tab:CIFAR100}
\end{table}

\begin{table}[!ht]
\centering
\caption{Comparison of deep active learning methods for SVHN dataset}
\label{tab:SVHN_candidate_time}
\footnotesize
\resizebox{\textwidth}{!}{
\begin{tabular}{|c|p{75mm}|c|c|c|c|}
\hline
\textbf{Model} & \textbf{Method} & \textbf{Candidate Model Used?} & \textbf{Time Saved (hrs)} & \textbf{Annotation Simulation Time (hrs)} & \textbf{Accuracy (\%)} \\ \hline

\multirow{5}{*}{DenseNet-121} 
& ALFA-Mix (large budget) \cite{parvaneh2022active} & Yes & 0.00 & - & 90.00 \\
& CUCB \cite{zhang2022nested} & Yes &  0.00 & - & 94.00 \\

&  Glister \cite{killamsetty2021glister} & Yes & 0.00 & 15.50 & $91.26 \pm 0.11$ \\

&  \textbf{BADGE \cite{ash2019deep}} & \textbf{Yes} & \textbf{0.00} & \textbf{25.92} & \textbf{$95.80 \pm 0.07$} \\

&  BAIT \cite{ash2021gone} & Yes & 0.00 & 30.06 & $91.37 \pm 0.14$ \\

& Ours (LC) & No & 0.94 & 16.96 & 95.77 $\pm$ 0.01 \\
& Ours (HC) & No & 0.94 & 16.96 & 95.48 $\pm$ 0.15 \\
& Ours (HCLC) & No & 0.94 & 16.96 & 95.76 $\pm$ 0.0 \\ \hline

\multirow{5}{*}{ResNet-56} 
& Double-Cross Attacks \cite{vicarte2021double} & Yes & 0.00 & - & 67.19 \\
& \textbf{Linear Feature Disentanglement \cite{he2022exploring}} & \textbf{No} & \textbf{-} & \textbf{-} & \textbf{96.66} \\
& Ours (LC) & No & 0.35 & 3.90 & 96.12 $\pm$ 0.11 \\
& Ours (HC) & No & 0.35 & 3.90 & 95.99 $\pm$ 0.05 \\
& Ours (HCLC) & No & 0.35 & 3.90  & 96.12 $\pm$ 0.11 \\ \hline

\multirow{9}{*}{VGG-16} 
& Pretext-based AL \cite{bhatnagar2020pal} & Yes & 0.00 & - & 94.00 \\
& BADGE  \cite{ash2019deep} (50K) & Yes & 0.00 & & 92.00 \\
& Ours (LC) (50K) & No & 0.27 & 3.72 & 94.22 \\
& \textbf{RL based AL \cite{sun2019active}} & \textbf{Yes} & \textbf{0.00} & \textbf{-} & \textbf{97.00} \\
& BAL at Scale \cite{citovsky2021batch} & Yes & 0.00 & - & 93.00 \\
& Co-Auxiliary Learning \cite{wang2023active} & Yes & 0.00 & - & 96.00 \\
& Ours (LC) & No & 0.27 & 4.39 & 95.51 $\pm$ 0.08 \\
& Ours (HC) & No & 0.27 & 4.39 & 95.45 $\pm$ 0.07 \\
& Ours (HCLC) & No & 0.27 & 4.39 & 95.61 $\pm$ 0.09 \\ \hline

\multirow{12}{*}{ResNet-18} 
& \cite{beck2021effective} & Yes & 0.00 & - & 95.10 \\
& ActiveLossNet \cite{xie2023effective} (10K)  & Yes & 0.00 & - & 91.30 \\

& BAIT \cite{ash2021gone} (15K) & Yes & 0.00 & - & 75.00 \\
& Ours (LC) (15K) & No & 0.29 & 0.46 & 91.8 $\pm$ 0.06 \\

& BADGE \cite{ash2019deep} (50K)  & Yes & 0.00 & - & 91.00 \\
& Ours (LC) (50K) & No & 0.29 & 2.06 & 93.23 $\pm$ 0.30 \\
& Random (Baseline) (10K) & No & 0.00 &  0.19 & 90.50 \\
& VAAL \cite{sinha2019vaal} (10K) & Yes & 0.00 & - & 92.00 \\
& Learning Loss \cite{wu2019learning} (10K) & Yes & 0.00 & - & 92.50 \\

& CoreGCN \cite{caramalau2021sequential} (10K) & Yes & 0.00 & - & 92.50 \\

& UncertainGCN  \cite{caramalau2021sequential} (10K) & Yes & 0.00 &  - & 92.20 \\

& Coreset  \cite{sener2018active} (10K) & Yes & 0.00 & - & 93.00 \\
& FeatProp  \cite{ye2016spatial} (10K) & Yes & 0.00 & - & 90.50 \\
& \textbf{Ours (LC)} & \textbf{No} & \textbf{0.29} & \textbf{2.84} & \textbf{95.84 $\pm$ 0.09} \\
& Ours (HC) & No & 0.29 & 2.84 & 95.65 $\pm$ 0.08 \\
& Ours (HCLC) & No & 0.29 & 2.84 & 95.83 $\pm$ 0.02 \\ \hline

\end{tabular}
}
\begin{minipage}{\linewidth}
\footnotesize 
\tiny \textit{Note:} DenseNet-121 results for BADGE, BAIT, and GLISTER (highlighted in blue) are reproduced using official code-bases under the same experimental conditions (SVHN, DenseNet-121, 3 runs). All other baseline results are taken from their respective original publications.
\end{minipage}
\label{tab:SVHN}
\end{table}

\begin{figure}[!ht]
    \centering
    \includegraphics[width=0.85\textwidth]{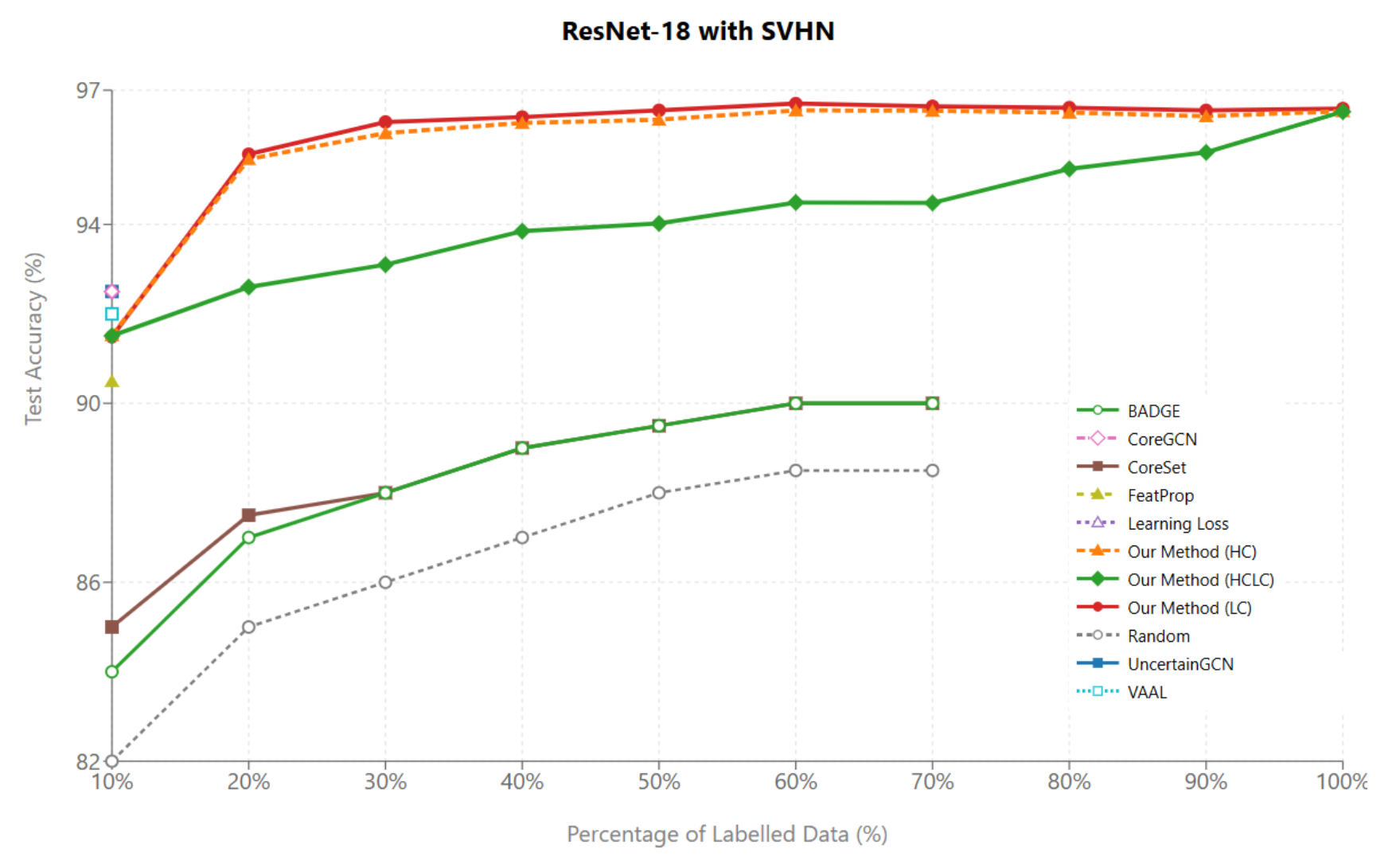}
    \caption{ 
    Comparison of active learning methods on the SVHN dataset with ResNet-18. 
    Baselines include Random, Coreset \cite{sener2018active}, 
    VAAL \cite{sinha2019vaal}, Learning Loss \cite{wu2019learning}, 
    FeatProp \cite{ye2016spatial}, CoreGCN \cite{caramalau2021sequential}, and BADGE \cite{ash2019deep}.}
    \label{fig:svhn_plot}
\end{figure}

The SVHN dataset enables us to test the candidate-model-free active learning methods on another domain of images. The obtained active learning results using DenseNet-121 are compared with state-of-the-art AL methods such as ALFA-Mix \cite{parvaneh2022active}, CUCB \cite{zhang2022nested} as shown in Table \ref{tab:SVHN}. DenseNet-121 achieved an accuracy of $\approx$ 95.77\% using both LC, HCLC methods, which outperformed ALFA-Mix (90\%) and CUCB \cite{zhang2022nested} (94\%). Similar to CIFAR-10, CIFAR-100, we have reproduced BADGE, GLISTER, and BAIT active learning results for SVHN using DenseNet-121. BADGE demonstrated strong stability with an accuracy of $95.80\% \pm 0.07\%$, while GLISTER ($91.26\% \pm 0.11\%$) and BAIT ($91.37\% \pm 0.14\%$) provided consistent performance metrics, establishing a clear baseline for our proposed strategy.
ResNet-56 obtained 96.12\%, a comparable accuracy with both LC, HCLC methods, compared to Linear Feature Disentanglement (96.66\%). However, observed dramatic improvements compared to Double-cross Attacks (67.19) \cite{vicarte2021double}.
VGG-16 demonstrated similar improvements, with HCLC achieving 95.61\%, outperforming pretext-based active learning (94\%) \cite{bhatnagar2020pal}, BAL at Scale \cite{citovsky2021batch} (93\%)  and showcasing comparable results with reinforcement learning based active learning (97\%) \cite{sun2019active} and Co-Auxillary Learning \cite{wang2023active} (96\%). At the 50K budget setting, LC method achieves 94.22\%, which is 2.22\% improvement compared to BADGE (50K) \cite{ash2019deep} (92\%). ResNet-18 also achieved potential improvements, with our LC approach reaching 95.84\% accuracy, outperforming traditional methods such as \cite{beck2021effective} (95.10\%). At the 15K budget setting, our method (91.80\%) shown dramatic improvements compared to BAIT \cite{ash2021gone}. However, other methods such as VAAL \cite{sinha2019vaal} (92\%), Coreset \cite{sener2018active} shown improved results at 10K budget settings due to their dependency on the candidate models. Our methods achieve superior performance compared to VAAL  \cite{sinha2019vaal} (92.00\%), FeatProp (91.00\%) \cite{ye2016spatial}, BADGE \cite{ash2019deep} (95\%),  and significantly exceeding the random baseline (90.50\%) as shown in Figure \ref{fig:svhn_plot}.

The performance gains observed over the SVHN dataset further demonstrate the robustness and generalizability of the sampling methods that do not relay on any candidate models, particularly in real-world high-variability scenarios. These results show that the proposed strategies effectively balance the exploration and improve model performance across different architectures and datasets. Another benefit of the proposed approach is efficiency. Traditional active learning requires repeated training of a candidate model to choose samples for further model improvement in the subsequent iterations. Our approach bypasses this, saving significant computation (especially in large-budget settings). We have conducted a few experiments using ResNet-18 on the SVHN dataset to empirically observe the time savings achieved with the proposed approach. The conventional active learning approach requires an additional ~ 0.29 hours (18 minutes) time to obtain a candidate model. The time taken to get a candidate model will increase exponentially if we have to train the larger neural networks.

\begin{table}[!ht]
\centering
\caption{Comparison of deep active learning methods for TinyImageNet dataset (ResNet-18)}
\label{tab:TinyImageNet_candidate_time}
\resizebox{\linewidth}{!}{
\begin{tabular}{|c|p{70mm}|c|c|c|c|}
\hline
\textbf{Model} & \textbf{Method} & \textbf{Candidate Model Used?} & \textbf{Time Saved (hrs)} & \textbf{Annotation Simulation Time (hrs)} & \textbf{Accuracy (\%)} \\ \hline

\multirow{16}{*}{ResNet-18} 
& \cite{sinha2021initial} & Yes & 0.00 & - & 35.00 \\
& \cite{hu2021reducing} & Yes & 0.00 & - & 52.00 \\
& \cite{yoo2018active} & Yes & 0.00 & - & 42.70 \\
& Random (baseline) (10K) & No & - & - & 34.50 \\

& VAAL \cite{sinha2019vaal} (10K) & Yes & 0.00 & - & 39.80 \\
& Learning Loss  \cite{wu2019learning} (10K) & Yes & 0.00  &  - & 42.10 \\
& Coreset  \cite{sener2018active} (50K) & Yes & 0.00 & - & 45.20 \\

& ALPF  \cite{hu2018active} (50K) & Yes & 0.00 & - & 42.70 \\

& FeatProp  \cite{ye2016spatial} (10K) & Yes & 0.00 & -  & 41.80 \\
& \cite{jung2023simple} & Yes & 0.00 & - & 39.79 \\
& \cite{huactive} & Yes & 0.00 & - & 44.10 \\
& \cite{bengar2021reducing} & Yes & 0.00 & - & 51.00 \\

& Progressive Active Learning\cite{yang2023not} & Yes & 0.00 & - & 52.10 \\

& \cite{bengar2022class} & Yes & 0.00 & - & 56.00 \\

& DQAS \cite{zhao2024dataset} & Yes & 0.00 & - & 53.42 \\

& BUAL \cite{zong2024bidirectional} & Yes & 0.00 & - & 46.50 \\

& \textbf{BADGE \cite{ash2019deep}} & \textbf{Yes} & \textbf{0.00} & \textbf{44.66} & \textbf{56.73 $\pm$ 0.12} \\
& BAIT \cite{ash2021gone} & Yes & 0.00 & 53.92 & 34.13 $\pm$ 0.03 \\
& GLISTER \cite{killamsetty2021glister} & Yes & 0.00 & 27.21 & 50.85 $\pm$ 0.02 \\
& SSAL \cite{li2022empirical} & Yes & 0.00 & - & 54.50 \\
& LC (Pre-trained) & Yes & 0.00 & 29.87 & 56.43 $\pm$ 0.12 \\
& Ours (LC) & No & 1.20 & 29.87 & 55.99 $\pm$ 0.12 \\
& Ours (HC) & No & 1.20 & 29.87 & 54.94 $\pm$ 0.14 \\
& Ours (HCLC) & No & 1.20 & 29.87 & 55.92 $\pm$ 0.11 \\ \hline

\end{tabular}
}
\begin{minipage}{\linewidth}
\footnotesize 
\tiny \textit{Note:} Results for BADGE, BAIT, and GLISTER (highlighted in blue) are reproduced using official code-bases under the same experimental conditions (TinyImageNet, ResNet-18, 3 runs). All other baseline results are taken from their respective original publications.
\end{minipage}
\label{tab:TinyImageNet}
\end{table}

\begin{itemize}
    \item \textbf{VGG-16:} Ideal for scenarios where simplicity and architectural consistency are prioritized.
    \item \textbf{ResNet:} The go-to choice for complex tasks requiring deep networks with high accuracy.
\end{itemize}

The selection of an appropriate architecture depends on the specific requirements of the application, the available computational resources, and the desired trade-offs between accuracy, efficiency, and ease of implementation. 

To verify the efficacy of the proposed method over higher resolution image datasets, active learning experiments are conducted with ResNet-18 using TinyImageNet. Table \ref{tab:TinyImageNet} presents the comparison of state-of-the-art active learning with the proposed method. From Table \ref{tab:TinyImageNet}, it is evident that the proposed active learning framework outperforming the other methods. Moreover, the difference between pre-trained ResNet-18 (ImageNet) and randomly initialized ResNet-18 on TinyImageNet is $<$ 1\%. Here, a very important thing to note is that the target task is similar to the source task. For many practical scenarios, the performance gap between pre-trained and random-initialization approaches is small enough that the computational savings justify skipping candidate model training. From this results, we can also infer that methods such BADGE \cite{ash2019deep} (56.73\%) achived marginal improvement compared to our methods. However, BADGE experiment took $\approx$45 hours time, which is almost $1.5\times$ compared with our methods. On the other hand, our methods outperform state-of-the-art active learning methods such as BAIT \cite{ash2021gone} (34.13\%), GLISTER \cite{killamsetty2021glister} (50.85\%), \cite{hu2021reducing} (52\%), and SSAL \cite{li2022empirical} (54.50\%). Also, our candidate free models outperformed compared to the recent methods DQAS \cite{zhao2024dataset} ($53.42\%$), BUAL \cite{zong2024bidirectional} ($46.50\%$), and Progressive Active Learning \cite{yang2023not} ($52.10\%$).

\begin{table}[!ht]
\centering
\caption{Comparison of SSD-based object detection methods on PASCAL VOC 2012}
\label{tab:PASCAL_VOC_candidate_time}
\footnotesize
\resizebox{\textwidth}{!}{
\begin{tabular}{|l|p{40mm}|c|c|c|c|}
\hline
\textbf{Method} & \textbf{SSD Variant / Backbone} & \textbf{Candidate Model Used?} & \textbf{Time Saved (hrs)} & \textbf{Annotation Simulation Time (hrs)} & \textbf{mAP} \\ \hline

SSD-300 \cite{liu2016ssd}
& SSD-300 / VGG-16
& No
& - & - & 73.1 \\ \hline

\textbf{FSSD-300 \cite{li2017fssd}}
& \textbf{FSSD-300 / VGG-16}
& \textbf{Yes}
& \textbf{0.00} & \textbf{-} & \textbf{82.7} \\ \hline

FF-SSD-300/512 \cite{zhao2017feature}
& FF-SSD-300 / 512 / VGG-16
& No
& - & - & 80.8 \\ \hline

Concat-SSD-300/512 \cite{zhao2017concatenated}
& Concat-SSD-300 / 512 / VGG-16
& No
& - & - & 80.8 \\ \hline

PPAL \cite{zhang2024ppal}
& ISD-SSD / VGG-16
& Yes
& 0.00 & - & 74.0 \\ \hline

Ours (LC)
& SSD / VGG-16
& No
& 2.58 & 9.26 &  $81.53 \pm 0.09$ \\ \hline

Ours (HC)
& SSD / VGG-16
& No
& 2.58 & 9.26 & $80.97 \pm 0.21 $ \\ \hline

Ours (HCLC)
& SSD / VGG-16
& No
& 2.58 & 9.26 & $78.13 \pm 0.09$ \\ \hline

\end{tabular}
}
\label{tab:PASCAL_VOC}
\end{table}

The PASCAL VOC 2012 dataset provides a strong benchmark for evaluating the effectiveness of our proposed active learning strategies in object detection. In Table \ref{tab:PASCAL_VOC}, comparison among the various SSD-based baselines and competitive approaches like PPAL \cite{zhang2024ppal} and FSSD \cite{li2017fssd} is presented. The proposed methods consistently delivered high performance. Our LC strategy achieved the highest reported mAP of 81.53\%, outperforming both PPAL (74\%) and established baselines like FF-SSD-300/512 (80.8\%). The HC variant followed closely with 80.97\%, while HCLC also showed solid performance at 78.13\%. our LC variant reached a strong mAP of 81.53\%. These results reflect the strength of our sampling strategies in selecting informative examples that improve model performance with fewer annotations. In particular, LC stood out for its steady gains across iterations, suggesting stronger generalization and stability. Overall, the improvements across all four variants highlight the practical effectiveness of our approach, especially in challenging real-world detection scenarios where annotation cost and data diversity are key concerns.

\section{Ablation Study}
There are a number of experiments being conducted to explore the impact of using a combination of different acquisition functions in active learning and to determine the best approach to use in choosing training samples for deep learning models. Active learning usually relies on the uncertainty of the model to label informative samples; however, how the acquisition functions are combined can significantly affect the learning process. Sampling approaches such as Low Confidence + High Confidence (LCHC) and Random + Low Confidence (RLC) are being explored to examine possible trade-offs between exploration and exploitation. This will help to understand whether a combination of both approaches can speed up the learning process or whether the emphasis on one type of sample is more beneficial. Moreover, the impact of randomness in the sampling process is also being investigated. By using random sampling in combination with confidence-based approaches, namely Random + High Confidence (RHC), it will be possible to determine whether the addition of diversity in the sampling process can improve the robustness of the model. Of particular interest is the combination approach of Hybrid Least Confidence and High Confidence (HLH), which provides a balanced approach that can potentially leverage the benefits of both approaches.

\begin{figure}[!tbp]
  \centering
  \begin{minipage}[b]{\textwidth}
    \caption{Comparison of various developed sampling methods results on CIFAR-10 using DenseNet-121.}
    \label{fig:ablation-resnet}
    \includegraphics[width=\textwidth]{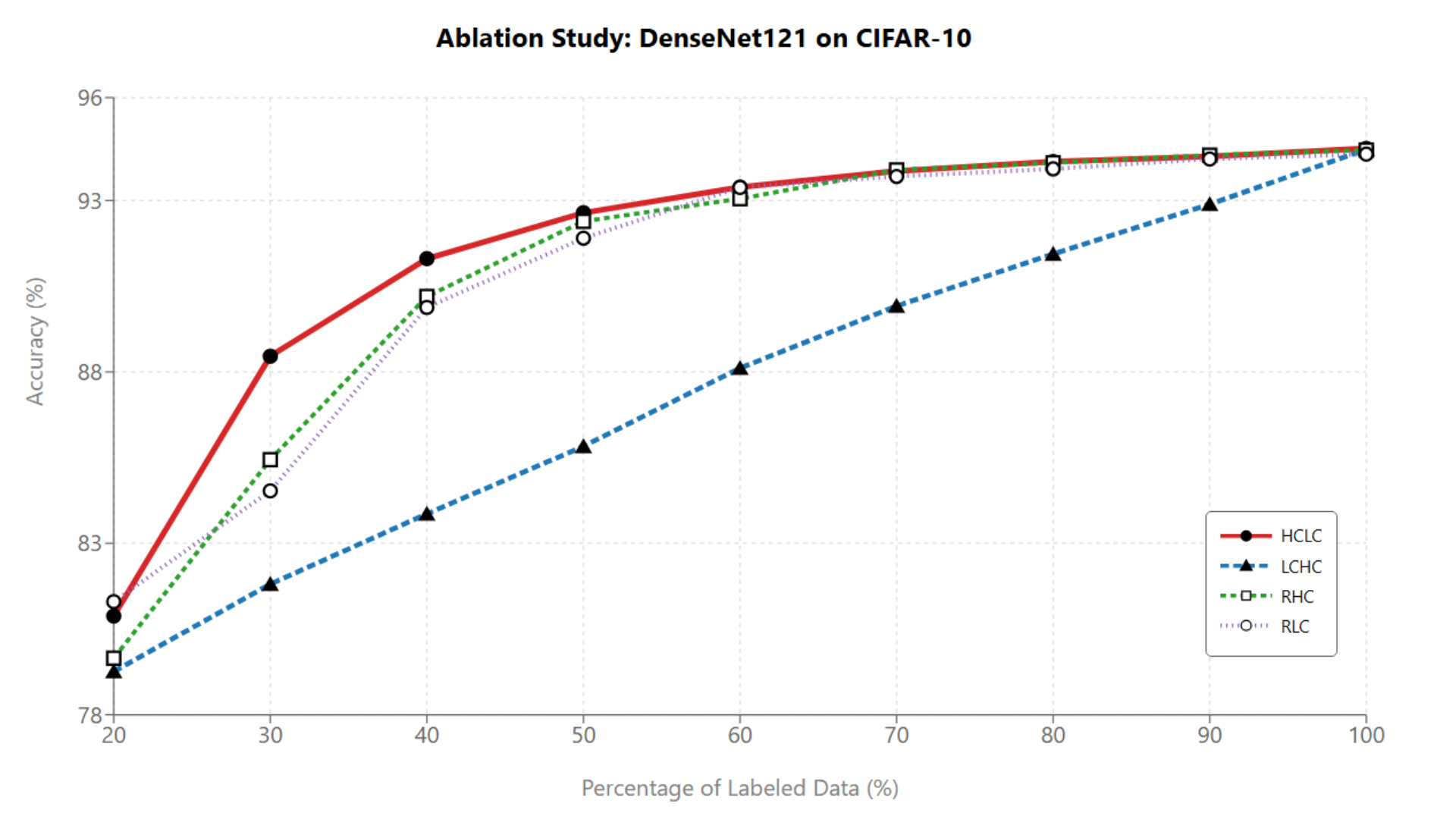}
    
  \end{minipage}
\end{figure}

\begin{table}[!ht ]
\caption{Results of various sampling methods developed to verify the impact of deep active learning on CIFAR-10.}
\label{tab:Ablation Study}
\begin{tabular}{|l|l|l|}
\hline
Model                & Methods                                                        & Accuracy                                                              \\ \hline
DenseNet-121 \cite{huang2017densely}         & \begin{tabular}[c]{@{}l@{}}LCHC\\ \textbf{HLH}\\ RHC\\ RLC\end{tabular} & \begin{tabular}[c]{@{}l@{}}94.49\\ \textbf{94.52}\\ 94.47\\ 94.35\end{tabular} \\ \hline
ResNet-56 \cite{he2016deep}           & \begin{tabular}[c]{@{}l@{}}LCHC\\ \textbf{HLH}\\ RHC\\ RLC\end{tabular} & \begin{tabular}[c]{@{}l@{}}92.99\\ \textbf{93.41}\\ 91.89\\ 92.30\end{tabular} \\ \hline
VGG-16  \cite{simonyan2014very}             & \begin{tabular}[c]{@{}l@{}}LCHC\\ \textbf{HLH}\\ RHC\\ RLC\end{tabular} & \begin{tabular}[c]{@{}l@{}}93.97\\ \textbf{94.35}\\ 94.05\\ 94.17\end{tabular} \\ \hline
ResNet-18 \cite{he2016deep}     
& \begin{tabular}[c]{@{}l@{}}LCHC\\ HLH\\ RHC\\ \textbf{RLC}\end{tabular} & \begin{tabular}[c]{@{}l@{}}95.24\\ 95.62\\ 95.24\\ \textbf{95.64}\end{tabular} \\ \hline
MoblienetV2 \cite{howard2017mobilenets} & \begin{tabular}[c]{@{}l@{}}LCHC\\ HLH\\ RHC\\ \textbf{RLC}\end{tabular} & \begin{tabular}[c]{@{}l@{}}94.67\\ 95.60\\ 94.69\\ \textbf{95.61}\end{tabular} \\ \hline
\end{tabular}
\end{table}

The results of the experiments is shown in Table \ref{tab:Ablation Study}, and it reveals some interesting observations regarding the impact of different combinations of acquisition functions on the performance of deep active learning. These experiments were performed on CIFAR-10, and the observations made are quite comprehensive in terms of different sampling methods. Among the different sampling methods, the hybrid acquisition function always performs better than other combinations of acquisition functions in terms of the final accuracy of the model. For example, for VGG-16, the hybrid method has an accuracy of 94.35\% at iteration 8, while LCHC has an accuracy of 93.97\%. ResNet-18 and MobileNet-V2 also demonstrate a significant improvement with the hybrid method, with ResNet-18 having an accuracy of 95.62\% (as shown in Fig. \ref{fig:ablation-resnet}) and MobileNet-V2 having an accuracy of 95.42\%.

By contrast, the random $+$ low (RLC) and random $+$ high (RHC) strategies have more variability between architectures. While the RLC strategy does see improvements for some architectures, such as MobileNetV2 and ResNet-18, its performance has largely plateaued or not reached the level of improvement seen in the hybrid strategy. This indicates that while random selection can add diversity to the selection method, it is not necessarily better than more focused strategies. Furthermore, the HC combinations, which rely on high-confidence samples for both the 10k and 5k selections, demonstrate slower accuracy improvements and underperform relative to the hybrid and other strategies, particularly in VGG-16 and ResNet-56. This suggests that an over-reliance on high-confidence samples may limit the model’s ability to explore diverse regions of the input space, thus impeding optimal learning. Overall, the results indicate that the hybrid acquisition function strategy provides the most effective approach for deep active learning among the four strategies explored in this ablation study, facilitating better model convergence and higher final accuracy across a range of architectures.

\begin{figure}[!tbp]
  \centering
  \begin{minipage}[b]{\textwidth}
    \caption{Comparison of active learning results for DinoV2 using pre-trained vs. DinoV2 with random weights on LabelMe1250K dataset.}
    \label{fig:dinov2_results}
    \includegraphics[width=\textwidth]{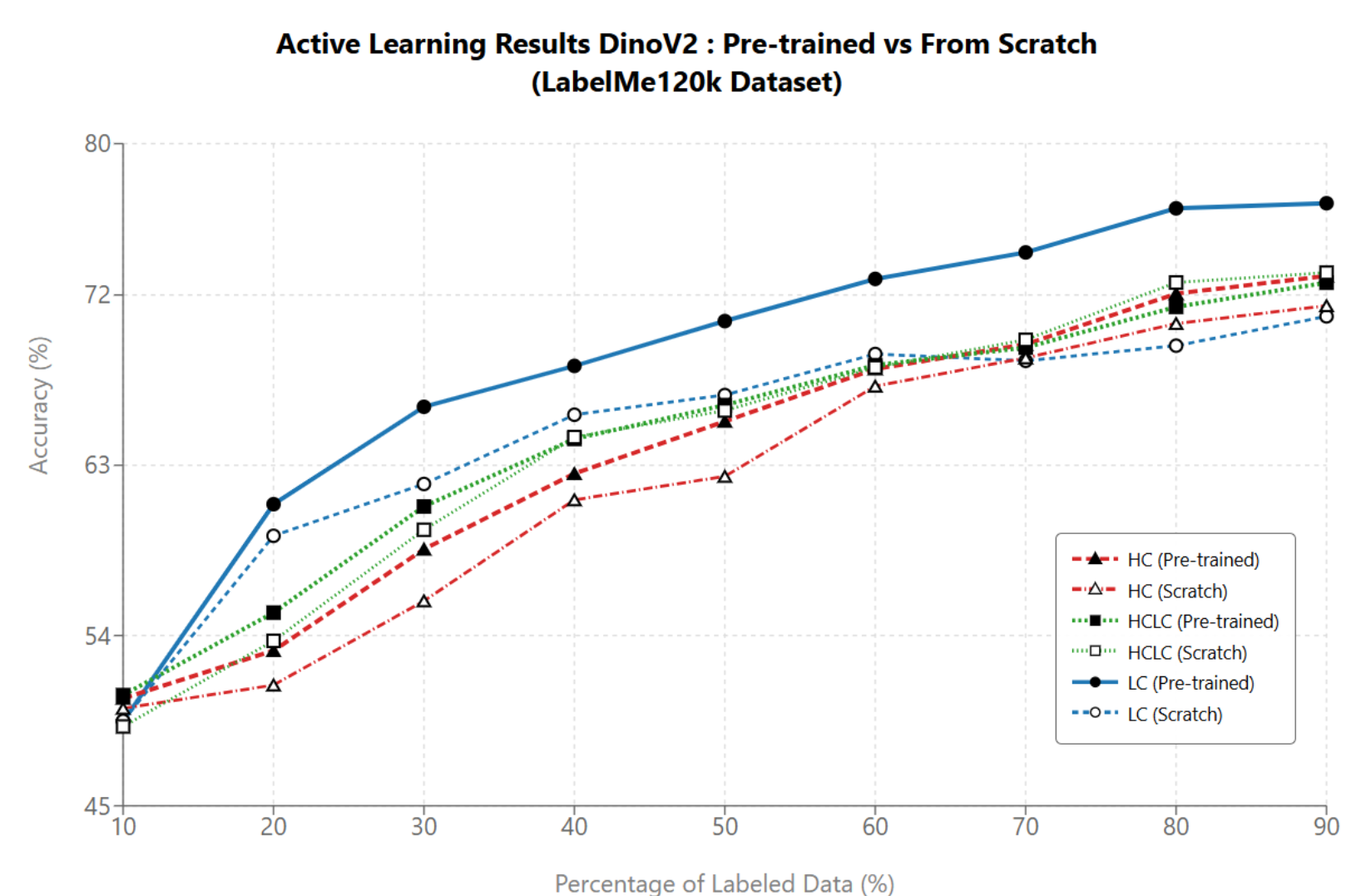}
    
  \end{minipage}
\end{figure}

We have verified the robustness of the proposed method using foundation model DinoV2 \cite{oquab2024dinov2}. Two active learning experiments i) finetuning the pre-trained DinoV2, ii). training DinoV2 model with random weights from scratch have been conducted. These experiments helped us to verify the claim of this paper when we have access to foundation model. The active learning results using DinoV2 on LabelMe1250K dataset \cite{uetz2009large} are presented in Fig. \ref{fig:dinov2_results}. It is evident from Fig. \ref{fig:dinov2_results} that pre-trained DinoV2 is exhibiting relatively better results compared to DinoV2 model when trained from scratch using the samples pooled using the proposed active learning methods. On the other hand, when DinoV2 is trained from scratch using the samples pooled using the hybrid sampling method, comparable results are observed.  Therefore, the proposed sampling methods yield consistent improvements in both the scratch and foundation‑model settings, and can be readily combined with strong pre‑trained backbones in practice. 

\subsection{Limitations under Class Imbalance}
To verify the limitations of candidate-free active learning, we have created an imbalanced CIFAR-10 variant with a 10:1 imbalance ratio, where: i) Class 1 (majority): all 5,000 training samples, ii) Class 10 (minority): only 500 training samples (10\% of the majority). Through this class imbalance experiment, we observe that 92.95\%, 93.85\%, and 92.50\% accuracies using LC, HCLC, and HC sampling methods, respectively. Our candidate-model-free approach demonstrates resilience but not immunity to extreme class imbalance. The performance gap between LC and HCLC highlights that the choice of acquisition function matters significantly when data distributions are highly skewed. In such cases, a hybrid strategy (HCLC) is preferable over pure uncertainty sampling.

\subsection{Computational Efficiency and Fair Comparison}

All experiments are conducted under controlled conditions using an identical hardware (NVIDIA GeForce RTX 4080 GPU with 16GB memory). Two key timing components were measured: i) \textbf{Time Saved:} The initial training period required for candidate-model-based methods before active selection can begin. This corresponds to training a model on the initial labeled set until convergence. ii). \textbf{Annotation Simulation Time:} The cumulative time spent across all active learning iterations, including sample selection and model updates after each annotation batch. Our candidate-model free methods bypass the initial training phase due to which there is reduction in training time along with annotation simulation time, which are reported in Tables \ref{tab:CIFAR10}, \ref{tab:CIFAR100}, \ref{tab:SVHN}, \ref{tab:TinyImageNet}, and \ref{tab:PASCAL_VOC}.

\section{Conclusion}
In this paper, we explored an important question, "Are candidate models really needed for deep active learning?". Our experiments demonstrated that models without pre-trained candidate models can still achieve competitive results through carefully designed sampling strategies. Specifically, we evaluated high confidence (HC), low confidence (LC), and a combination of both, high confidence and low confidence (HCLC) sampling methods, with LC emerging as the most effective strategy across multiple datasets and architectures. By eliminating the reliance on candidate models, our approach simplifies the active learning pipeline, making it more accessible and scalable across different domains.

While our study highlights the advantages of this approach, there are several avenues for future exploration. Extending our strategies to more complex datasets and real-world applications could provide deeper insights into their generalizability. Further research could investigate the impact of choosing sub-networks instead of the full network for sample selection.

	\section*{Credit authorship contribution statement}
	\textbf{First Author:} Conceptualization, Methodology, Software, Writing – initial draft \& editing. \textbf{Second Author:} Methodology, Software, Writing – initial draft \& editing. \textbf{Third Author:} Methodology, Software, Writing. \textbf{Fourth Author:} Supervision, Methodology, Validation, Writing – editing and reviewing.
    \textbf{Last Author:} Methodology, Software, Writing – editing.

	\section*{Declaration of Competing Interest}
	The authors declare that they have no known competing financial interests or personal relationships that could have appeared to influence the work reported in this paper.
	
	\section*{Acknowledgements}
	We are grateful to VectraTech Global for providing us the three NVIDIA GeForce Titan X Pascal 12GB GPU cards that are used for this research.
	
	\section*{Data availability}
	Provide a link to your data if available.
    
    \bibliographystyle{model5-names}\biboptions{authoryear}
	\bibliography{egbib}
	
\end{document}